\documentclass[runningheads]{llncs}

\usepackage{accv}

\usepackage{accvabbrv}

\usepackage{graphicx}
\usepackage{booktabs}

\usepackage[accsupp]{axessibility}  

\usepackage[pagebackref,breaklinks,colorlinks,citecolor=accvblue]{hyperref}

\usepackage{orcidlink}

\begin{document}
\newsavebox\CBox
\def\textBF#1{\sbox\CBox{#1}\resizebox{\wd\CBox}{\ht\CBox}{\textbf{#1}}}
\title{Towards Reflected Object Detection: A Benchmark}

\titlerunning{Towards Reflected Object Detection: A Benchmark}

\author{Yiquan Wu\inst{1,}\and Zhongtian Wang\inst{1,}\and You Wu\inst{2,}\and
Ling Huang\inst{2,}\and
Hui Zhou\inst{2}\and
Shuiwang Li\inst{2}$*$}

\authorrunning{Z.Wang~ et al.}

\institute{Nanjing University of Aeronautics and Astronautics, China\and
Guilin University of Technology, China \\
\email{lishuiwang0721@163.com}\\
}

\maketitle

\begin{abstract}

Object detection has greatly improved over the past decade thanks to advances in deep learning and large-scale datasets.
However, detecting objects reflected in surfaces remains an underexplored area. Reflective surfaces are ubiquitous in daily life, appearing in homes, offices, public spaces, and natural environments. Accurate detection and interpretation of reflected objects are essential for various applications.
This paper addresses this gap by introducing a extensive benchmark specifically designed for Reflected Object Detection. 
Our Reflected Object Detection Dataset (RODD) features a diverse collection of images showcasing reflected objects in various contexts, providing standard annotations for both real and reflected objects. This distinguishes it from traditional object detection benchmarks. RODD encompasses 10 categories and includes 21,059 images of real and reflected objects across different backgrounds, complete with standard bounding box annotations and the classification of objects as real or reflected.
Additionally, we present baseline results by adapting five state-of-the-art object detection models to address this challenging task. Experimental results underscore the limitations of existing methods when applied to reflected object detection, highlighting the need for specialized approaches. By releasing RODD, we aim to support and advance future research on detecting reflected objects.
Dataset and code are available at: \url{https://github.com/jirouvan/ROD}.

  \keywords{Reflected object detection \and Benchmark \and Object detection}
\end{abstract}

\footnotetext[0]{$*$ Corresponding author.}

\section{Introduction}
\label{sec:intro}

\begin{figure*}[t]
		\centering
		\begin{minipage}[t]{1\textwidth}
			\centering
			\includegraphics[width=1\textwidth]{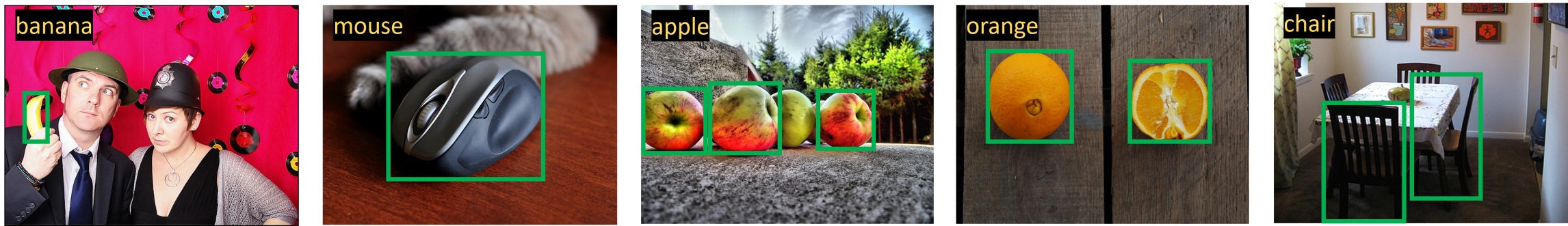}
			\centerline{(a) Example of conventional object detection.}
		\end{minipage}
		\begin{minipage}[t]{1\textwidth}
			\centering
			\includegraphics[width=1\textwidth]{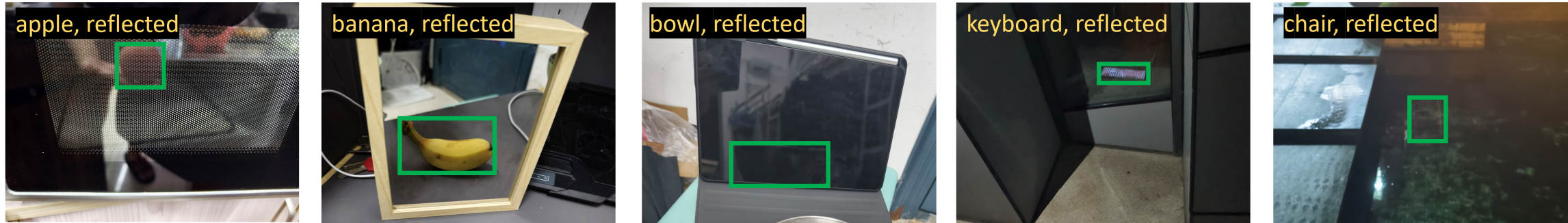}
			\centerline{(b) Example of detecting reflected objects.}
		\end{minipage}
		\caption{While previous object detection focused on the identification and localization of objects, this work focuses on information beyond that and concerns about the nature of objects in addition, as shown in (a) and (b), respectively. Note the nature of the objects (i.e., real or reflected) are marked in (b) additionally. 
		}
		\label{fig:RODD_examples1}\vspace{-0.3in}
	\end{figure*}

The field of object detection has seen remarkable advancements over the past decade, driven by the development of deep learning techniques and the availability of large-scale datasets \cite{Zhao2018ObjectDW,Zou2019ObjectDI,BenaliAmjoud2023ObjectDU}. These advancements have significantly improved the accuracy and robustness of object detection systems in various applications \cite{Liu2018DeepLF}. However, one area that remains underexplored is the detection of objects reflected in surfaces, such as mirrors, glass windows, and other reflective materials. See Fig. \ref{fig:RODD_examples1} for an illustration of
the difference between conventional object detection and reflected object detection.

Reflective surfaces are ubiquitous in our daily lives, appearing in a wide array of environments and applications \cite{Owen2019DetectingRB,Yang2019WhereIM,Lin2023LearningTD}. Mirrors, glass windows, water surfaces, and polished metals are just a few examples of materials that produce reflections. These reflective surfaces are prevalent in various settings, including homes, offices, public spaces, and natural environments, making the ability to detect and interpret reflected objects a crucial aspect of many technological applications \cite{Owen2019DetectingRB,Lin2020ProgressiveMD,Ji2021FullduplexSF,Liu2024MultiViewDR}.
For instance, in surveillance, security systems can more effectively identify real intrusions or threats by differentiating reflections from genuine objects \cite{Shen2018BlindSM,Singhal2022ECMSEC,Ray2022ThePB}. For autonomous driving, accurate identification of real objects versus reflections enables vehicles to navigate more safely and avoid accidents caused by misinterpretation \cite{noriaki2023collision,li2023traffic,zhang2022traffic,diaz2003optical}. For service robots, robots can perform tasks with greater accuracy, such as picking and placing items, by correctly identifying real objects instead of their reflections \cite{Park2021IdentifyingRI,Lin2023LearningTD}. This improved object detection also facilitates better navigation in environments with reflective surfaces, such as warehouses \cite{Lu2020ServiceRC,Damodaran2023ExperimentalAO}. In smart homes, systems can provide more tailored responses by recognizing when a person is truly present rather than reacting to their reflection \cite{marikyan2019systematic,sovacool2020smart,chakraborty2023smart}. In medical applications, imaging and diagnostic tools yield more accurate results when they accurately interpret reflections, leading to better patient outcomes and more precise medical interventions.

Given the widespread presence of reflective surfaces in daily life, developing technologies that can effectively detect and interpret reflected objects is essential. This capability can enhance the performance and reliability of various applications, including smart home systems, surveillance, autonomous driving, and medical devices. However, to the best of our knowledge, there is currently no public benchmark for reflected object detection. This paper aims to address this gap by introducing a benchmark specifically designed for this purpose. We propose a comprehensive benchmark that includes a diverse set of images featuring reflected objects in various contexts. Our benchmark is designed to test the limits of current object detection methods and provide a standardized evaluation framework for developing and comparing new algorithms tailored to reflected object detection. The benchmark provides standard annotations used in object detection for identifying both actual (real) objects and their reflections. Additionally, it offers extra details that indicate whether an object is real or a reflection. This feature distinguishes it from traditional object detection benchmarks, which typically do not provide information about whether an object is a reflection.
In addition to introducing the benchmark, this paper also presents baseline results by adapting several state-of-the-art object detection models. These results highlight the limitations of existing methods when applied to reflected object detection and underscore the need for specialized approaches. We analyze the performance of these models across different reflection scenarios and provide insights into the specific challenges posed by reflections.

\subsection{Contribution}

In this work, we make the first attempt to explore reflected object detection by introducing the RODD benchmark, which is specifically designed for detecting reflected objects. This benchmark provides a well-annotated dataset and robust evaluation metrics to facilitate research in this challenging area. 
The RODD benchmark fills a crucial gap in current object detection methods by focusing on reflected objects. It aims to provide researchers with a valuable resource to develop and test algorithms that handle the complexities of reflected objects.
RODD comprises a diverse set of 10 classes of generic objects, totaling 21059 images annotated with axis-aligned bounding boxes, category labels, and object nature (real or reflected). Sample images from the RODD dataset are illustrated in Fig. \ref{fig_anno}. In addition, we developed five baseline detectors based on five state-of-the-art algorithms, namely RO-YOLOV8, RO-YOLOV10, RO-RTMDet, RO-YOLOX, and RO-PPYOLOE. These baselines serve to evaluate detectors' performance and provide benchmarks for future research on RODD.
In summary, our contributions include:
\begin{itemize}
        \item We make the first attempt to explore detecting reflected objects, a previously underexplored area in object detection. By focusing on this unique challenge, we hope to inspire further research and innovation in the detecting reflected objects.
  
        \item We introduce RODD, the first benchmark dedicated to detecting reflected objects, which consists of 10 classes of generic objects, with 21,059 images annotated with bounding boxes, object categories, and the nature of the objects. This dataset will enable detailed analysis and evaluation of algorithms developed for detecting reflected objects.
  
        \item To support further research on RODD, we develop five baseline detectors based on state-of-the-art models: RO-YOLOV8, RO-YOLOV10, RO-RTMDet, RO-YOLOX, and RO-PPYOLOE. These baseline models will provide initial performance metrics and serve as reference points for future studies. 
	\end{itemize}


\section{Related Work}
\subsection{Object Detection Algorithms}

Object detection has been a critical area of research in computer vision, significantly advancing over the past few decades. Traditional object detection methods relied heavily on handcrafted features and shallow learning techniques. The advent of deep learning has revolutionized this field, leading to the development of more robust and accurate algorithms. 
Modern object detection methods are categorized into two types: two-stage detectors and one-stage detectors. Two-stage detectors, such as R-CNN \cite{girshick2014rich}, Fast R-CNN \cite{girshick2015fast}, Faster R-CNN \cite{ren2015faster}, and Mask R-CNN \cite{he2017mask}, initially generate region proposals and then refine them through classification and bounding box regression, achieving high precision and efficiency. Variants like Cascade R-CNN \cite{cai2018cascade} further enhance detection performance through multi-stage detection and regression. One-stage detectors, including SSD \cite{liu2016ssd}, YOLO, RetinaNet \cite{lin2017focal}, and EfficientDet \cite{tan2020efficientdet}, predict object locations and categories in a single step, providing faster performance suitable for real-time applications. The YOLO series has evolved to YOLOv8 \cite{yolov82023} and YOLOv10 \cite{wang2024yolov10}, further optimizing speed and accuracy.

Despite these advancements, detecting objects reflected in surfaces such as mirrors and glass remains a challenging and underexplored problem. Most existing object detection algorithms are not specifically designed to differentiate between real objects and their reflections, leading to potential false positives and degraded performance in environments with reflective surfaces. Our work aims to address this gap by introducing a benchmark and developing specialized approaches for reflected object detection.

\subsection{Object Detection Benchmarks}



Object detection benchmarks play a crucial role in the development and evaluation of detection algorithms by providing standardized datasets and evaluation metrics that facilitate consistent and fair comparisons among different approaches. Over the years, several prominent benchmarks have emerged, each contributing uniquely to the field, such as PASCAL VOC\cite{pascalvoc}, MS COCO \cite{lin2014microsoft}, and ImageNet \cite{krizhevsky2012imagenet}. These benchmarks provide large-scale images and standardized evaluation metrics. For instance, PASCAL VOC comprises 20 categories with 11,530 images and 27,450 annotated bounding boxes. ImageNet covers 200 categories with approximately 500,000 annotated bounding boxes. MS COCO includes 91 categories, over 300,000 images, and 2.5 million annotated instances. These datasets have been instrumental in pushing the boundaries of object detection research, promoting the development of more accurate and robust models. In addition to these established benchmarks, several domain-specific benchmarks have emerged to address particular challenges in object detection. For instance, KITTI \cite{Geiger2012CVPR} focuses on autonomous driving scenarios, providing annotated data for detecting objects such as cars, pedestrians, and cyclists in street scenes. UAVDT (UAV Detection and Tracking) \cite{du2018unmanned} provides benchmarks for aerial object detection, emphasizing challenges unique to unmanned aerial vehicle (UAV) imagery, such as varying altitudes and viewpoints.

Despite significant advancements in object detection, no public benchmark specifically targets reflected object detection. This gap hinders the development and evaluation of algorithms for handling reflections. Reflective surfaces are common in real-world scenarios such as surveillance, autonomous driving, and smart homes. Accurate detection of reflected objects is crucial for enhancing performance and safety in these applications. This paper addresses this gap by introducing a benchmark specifically tailored for reflected object detection.

\subsection{Dealing With Mirrors and Reflections in Vision}
Mirrors or other reflective surfaces are common in natural images, and can cause false positive results in the
tasks of detection, segmentation, counting, robotic navigation, scene reconstruction, and etc \cite{Bajcsy1996DetectionOD,DelPozo2007DetectingSS,Owen2019DetectingRB,Lin2020ProgressiveMD,Ji2021FullduplexSF,Liu2024MultiViewDR}. Reflection detection focuses on identifying regions in an image that contain reflections. When we take a picture through glass windows, the photographs are often degraded by undesired reflections.
One of the primary approaches to dealing with reflections involves removing or suppressing the reflections in images. For instance, Abiko et al. employed generative adversarial networks (GANs) to enhance the quality of reflection removal, yielding more natural and clear images \cite{Abiko2019SingleIR}. Arvanitopoulos et al. propose a single image reflection suppression method based on a Laplacian data fidelity and an l-zero gradient sparsity regularization term \cite{Arvanitopoulos2017SingleIR}.
Particularly, mirror surface detection aims to identify and segment mirror surfaces within a scene. For instance, Yang et al. proposed to address the mirror segmentation problem with a computational approach \cite{Yang2019WhereIM}. Since then, numerous methods have been developed to address mirror detection and segmentation \cite{Owen2019DetectingRB,Lin2020ProgressiveMD,Ji2021FullduplexSF,Liu2024MultiViewDR}.

Despite extensive research efforts dedicated to dealing with mirrors and reflections in vision, most of these works focus primarily on identifying, localizing, segmenting, and suppressing reflective regions in images. In this work, we make the first attempt to differentiate reflected objects from real ones, a critical capability for various applications, including surveillance, autonomous driving, service robots, and smart homes. 

\section{Benchmark for Reflected Object Detection}
We construct a dedicated dataset for Reflected Object Detection Dataset (RODD), which is a dataset that contains labels of both class and object nature, with prediction bounding-box labeled for each image.

\subsection{Image Collection}
\begin{figure}[t]
  \centering
  \includegraphics[width=12.3cm,height=7.6cm]{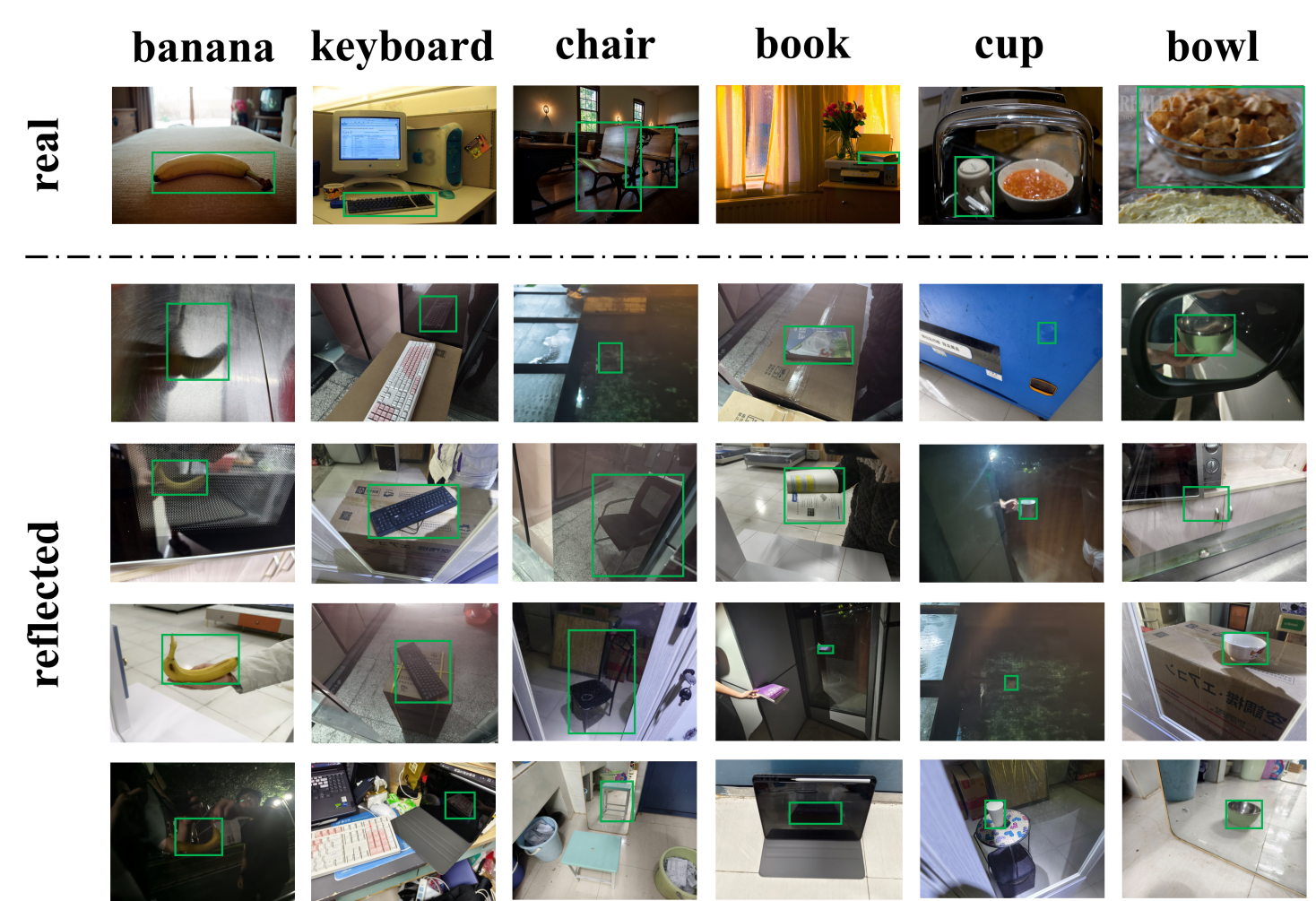}
  \caption{Samples from six categories (i.e., `banana', `keyboard', `chair',’book’, 'cup', and ’bowl’, from left to right) and their corresponding natures (i.e., `real' and `reflected' from top to bottom) in the RODD dataset. Note that the objects have been marked with green bounding boxes.}
  \label{fig_anno}
\end{figure}

For image collection, we selected 10 common objects in daily life, guided by the selection principles of PASCAL VOC \cite{everingham2010pascal} and COCO \cite{lin2014microsoft}. The chosen objects for RODD are bowl, apple, mouse, keyboard, banana, carrot, cup, orange, chair, and book, all of which are categories included in the COCO dataset. However, gathering varied images of these objects or their reflected ones in different scenes can be challenging. To address this, we initially sourced images using web crawlers and online repositories that focus on real-world scenarios with reflective surfaces. Additionally, we conducted field photography sessions in various environments such as homes, offices, and public spaces to capture images that include mirrors and other reflective surfaces. 
To ensure that the dataset was representative of real-world conditions, we made sure to capture images under various lighting conditions and from different angles. The final collection consists of 21,059 images, covering 10 objects (bowl, apple, mouse, keyboard, banana, carrot, cup, orange, chair, and book) and 2 attribute indicating the nature of the objects (i.e., real or reflected).
Fig. \ref{fig_anno} presents some sample images from RODD, demonstrating that each object category is captured in multiple scenes.

\noindent

\subsection{Annotation}

This section provides a detailed introduction to the image annotation process, covering three aspects: category, bounding box, and the nature of the object, as follows:

	\begin{itemize}
		\item \textBF{Category:} one of: bowl, apple, mouse, keyboard, banana, carrot, cup, orange, chair, and book.
		\item \textBF{Bounding box:} an axis-aligned bounding box that encloses the visible part of the object in the image.
        \item \textBF{Nature of the object:} a real or reflected object.
	\end{itemize}


\begin{figure}[t]
  \centering
  \includegraphics[width=0.95\linewidth]{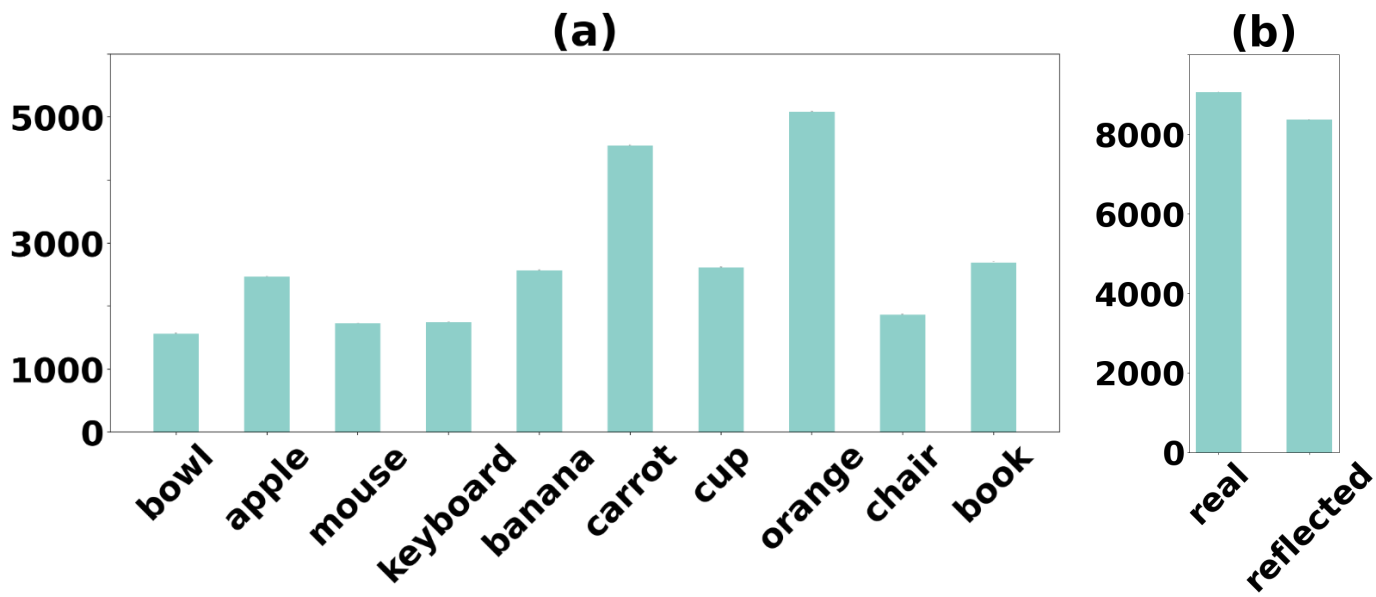}
  \caption{(a) Number of images per category in RODD.  (b) Number of images containing real or reflected objects in RODD.
  }
  \label{fig:statistics_of_DMSOD}
\end{figure}


We follow three steps, i.e., manual annotation, visual inspection, and box refinement, to complete the annotation of images, guided by the annotation guidelines proposed in \cite{everingham2010pascal} and \cite{lin2014microsoft}.
Specifically, all the images are first annotated by an expert, i.e., a student engaged in object detection, during the initial stage.
Manual annotation can lead to occasional errors or inconsistencies, prompting the verification team to carefully review the annotated files in the second step.
Annotation errors identified by the validation team in the third stage will be sent back to the initial annotation stage for refinement.
By employing this three-stage strategy, the dataset ensures its contained objects have high-quality annotation.
Fig. \ref{fig_anno} displays five examples of box annotations from RODD.

\subsection{Dataset Statistics}

The statistics of the RODD dataset are summarized in Fig. \ref{fig:statistics_of_DMSOD}. Fig. \ref{fig:statistics_of_DMSOD} (a) presents a histogram showing the number of images in the dataset for each category. As observed, the `orange' category is the most frequent, with 5,086 images. Fig. \ref{fig:statistics_of_DMSOD} (b) displays a histogram that illustrates the number of images containing real or reflected objects. This detailed breakdown highlights the distribution and prevalence of each object category within the dataset, providing insight into the dataset's composition and the representation of reflections. To facilitate training and evaluation, the RODD dataset is split into two primary subsets: the training set and the test set, with a ratio of 7:3.

\section{Baseline Detectors for Detecting Reflected Objects}

We develop five baseline detectors based on five state-of-the-art object detection algorithms, i.e., RTMDet \cite{lyu2022rtmdet}, YOLOv10 \cite{wang2024yolov10}, YOLOV8 \cite{yolov82023}, YOLOX \cite{ge2021yolox}, and PPYOLOE \cite{xu2022pp}, to facilitate the development of detecting reflected objects. For each model, we add an additional head or branch to predict the nature of the objects without altering the overall framework. The resulting baseline detectors are named RO-RTMDet, RO-YOLOv10, RO-YOLOV8, RO-YOLOX, and RO-PPYOLOE, respectively. Given space constraints and the fact that YOLOv10, YOLOV8, YOLOX, and PPYOLOE are all YOLO variants, we detail only RO-RTMDet and RO-YOLOv10 in the following sections. The extension to YOLOV8, YOLOX, and PPYOLOE is straightforward and will not be elaborated upon here.

\subsection{RO-RTMDet}

\begin{figure}[t]
  \centering
  \includegraphics[width=12cm,height=4cm]{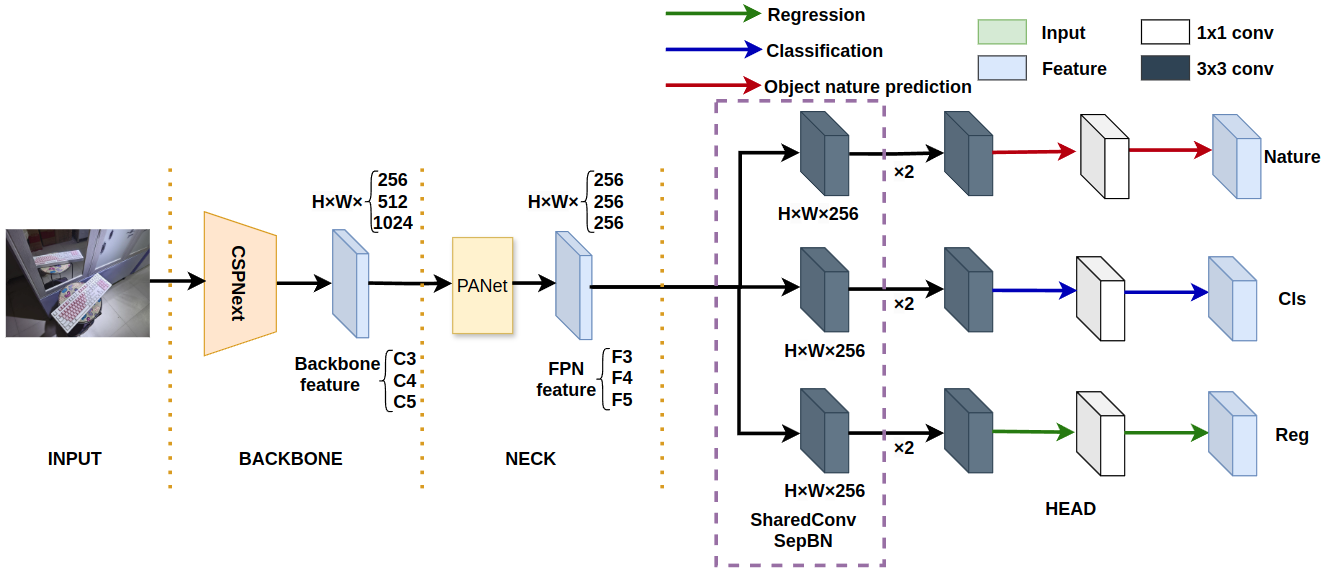}
  \caption{The network structure of the RO-RTMDet detector, inherited from RTMDet, is different from the addition of an additional branch head for the object nature.
  }
  \label{fig_RTMDet}
\end{figure}
The network architecture of the proposed RO-RTMDet is shown in  Fig. \ref{fig_RTMDet}.
CSPNet \cite{wang2020cspnet} serves as the backbone, generating output features C3, C4, and C5 with 128, 256, and 512 channels, respectively.
These features are fused into CSP-PAFPN \cite{lyu2022rtmdet}, the neck of RO-RTMDet, which employs the same block as the backbone.
The classification head and the regression head are two parallel components used for classification and regression, respectively, forming the head of the original RTMDet.
Building upon the original RTMDet model, we introduce a new classification head to predict the nature of objects (i.e., real or reflected).
During RO-RTMDet training, the overall loss of the model is defined as follows:
\begin{align}
L_{total} = L_{cls} + L_{reg} +\lambda L_{nat},
\end{align}
where $L_{cls}$, $L_{reg}$, and $L_{nat}$ represent the losses for classification, regression, and object nature prediction, respectively. $\lambda$ is a constant that weights the loss for the reflected objects prediction head.
Below are their specific definitions:
\begin{equation}
\small
\begin{aligned}
        & L_{cls} =\frac{1}{N_{pos}}\sum_{n=1}^{N_{pos}}\sum_{cls\in classes}-| y^{cls}_{n} - p^{cls}_{n} | ^\beta ((1 - y^{cls}_{n})log(1 -  p^{cls}_{n} ) + y^{cls}_{n} log( p^{cls}_{n})), \\
        & L_{reg} =\frac{1}{N_{pos}}\sum_{n=1}^{N_{pos}} 1 - (\text{IOU}(b^t_n,b^p_n) - \frac{|C-b^t_n\bigcup b^p_n|}{|C|}), \\
        & L_{nat} = \frac{1}{N_{pos}}\sum_{n=1}^{N_{pos}}\sum_{nat\in natures}-| y^{nat}_{n} - p^{nat}_{n} | ^\beta((1 - y^{nat}_{n})log(1 -  p^{nat}_{n} ) + y^{nat}_{n} log( p^{nat}_{n})),
\end{aligned}
\end{equation}
where $y^{cls}_{n}$ and $y^{nat}_{n}$ are the labeled value of classification and the object nature, $p^{cls}_{n} $ and $ p^{nat}_{n} $ are the corresponding predictions, $N_{pos}$ is the number of positive anchor,
$\beta$ is the hyperparameter for the dynamic scale factor, which is set to 2,
$b^t_n$ and $b^p_n$ represent the ground truth bounding boxes and the prediction, respectively;
$\text{IOU}$ and $C$ are the IOU loss function and the smallest
enclosing convex box of these two bounding boxes. We utilize the same training pipeline as RTMDet for training RO-RTMDet.

\begin{figure}[t]
  \centering
  \includegraphics[width=12cm,height=4cm]{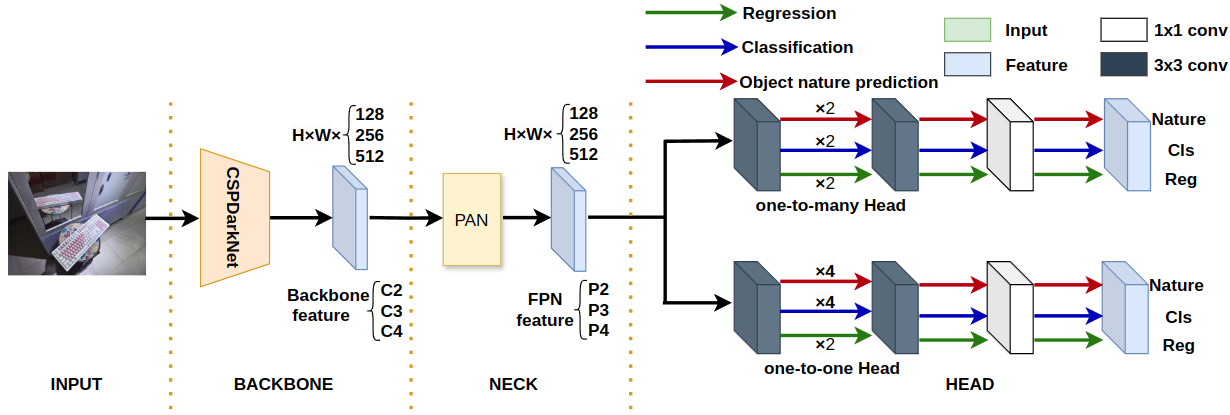}
  \caption{The network structure of the RO-YOLOv10 detector is inherited from YOLOv10, except for the addition of an additional reflected nature branch head.
  }
  \label{fig_yolov10}
\end{figure}
\subsection{RO-YOLOv10}

The network architecture of the proposed RO-YOLOv10 detector is shown in Fig. \ref{fig_yolov10}.
RO-YOLOv10 uses a modified CSPDarknet as backbone. It replaces the C2f module used in YOLOv8 \cite{yolov82023} with a compact inverted block (CIB) module and introduces an efficient partial self-attention (PSA) module \cite{wang2024yolov10}.
These features are input into the neck to enhance feature representation, which is made up of the PAN (Path Aggregation Network).
The original YOLOv10 model has two types of heads: (1) a one-to-many (o2m) head for regression and classification tasks, and (2) a one-to-one (o2o) head for precise localization.
In RO-YOLOv10, we add object nature prediction branch into both of these two heads.
During RO-YOLOv10 training, the overall loss of the model is defined as follows:

\begin{equation}
\small
\label{yolov10_loss}
\begin{aligned}
& L_{total} =  L_{o2m-head} + L_{o2o-head} , \\
&  L_{o2m-head}=  L_{o2m-cls } + \lambda L_{o2m-nat } + L_{o2m-reg } + L_{o2m-dfl }, \\ 
&   L_{o2o-head}=  L_{o2o-cls } + \lambda L_{o2o-nat } + L_{o2o-reg } + L_{o2o-dfl }
\end{aligned}
\end{equation}
In the o2m head, $L_{o2m-cls }$ and $L_{o2m-nat }$ represent the losses for classification and object nature prediction, respectively, while $L_{reg }$ and  $L_{dfl}$ indicate the CompleteIntersectin over Union (CIoU) Loss \cite{Zheng2020EnhancingGF} and  the Distribution Focal loss (DFL) \cite{Li2020GeneralizedFL}.
Similarly, each loss function in the o2o head carries the same meaning as in the the o2m head.
$\lambda$  is a constant that weights the loss for the object nature prediction branch.
Below, the $L_{cls }$ and $L_{nat }$ in the o2m head are used as examples to provide their specific definitions.
The specific definition of $L_{reg }$ and $L_{dfl}$ are omitted, as it is too intricate to elaborate on here and may divert from the main focus of our discussion. For a comprehensive understanding of $L_{reg }$ and  $L_{dfl}$, we recommend referring to the detailed explanations provided in the original documentation by Zheng et al. \cite{Zheng2020EnhancingGF} and Li et al. \cite{Li2020GeneralizedFL}.

\begin{equation}
\small
\begin{aligned}
    & L_{cls} = \frac{1}{N_{pos}}\sum_{n=1}^{N_{pos}}\sum_{cls\in classes}y_n^{cls}log(p_n^{cls})+(1-y_n^{cls})log(1-p_n^{cls}),\\
    & L_{nat} =\frac{1}{N_{pos}}\sum_{n=1}^{N_{pos}}\sum_{nat\in natures}y_n^{nat}log(p_n^{nat})+(1-y_n^{nat})log(1-p_n^{nat}),
\end{aligned}
\end{equation}
where $y^{cls}_{n}$ and $y^{nat}_{n}$ are the labeled value of classification and the object nature, $p^{cls}_{n} $ and $ p^{nat}_{n} $ are the corresponding predictions, $N_{pos}$ is the number of positive anchor. We utilize the same training pipeline as YOLOv10 for training RO-YOLOv10.


\section{Evaluation}
\subsection{Evaluation Metrics}
\label{section_metrics}
In the experiment, the proposed baseline detectors are evaluated for the performance by using two common metrics, i.e., average precision (AP) and mean average precision (mAP).
IOU (Intersection over Union) measures the overlap between the predicted bounding box (bbox) and the ground truth bbox.
In object detection tasks, a complete prediction comprises two main components: first, the model must identify specific objects within a given image, and second, it needs to accurately determine their respective locations.
Specifically, precision is the proportion of objects predicted by the model that match the real objects, whereas recall measures the proportion of real objects detected by the model.
These two measures are combined in mAP, which highlights the significance of properly balancing each during the evaluation process.

Guided by the COCO evaluation \cite{lin2014microsoft}, three IoU thresholds are used: fixed thresholds at 0.5 and 0.75 and a range threshold from 0.5 to 0.95 with a step size of 0.05. The corresponding average precisions (APs) are evaluated under these IoU thresholds, denoted as AP@0.5, AP@0.75, and AP@[.50:.05:.95], respectively.
In the experiment, COCO mAP is employed to evaluate the performance of detectors in detecting reflected objects.
Following \cite{qin2022detection,guo2022identity,wu2023liquid,li2023beyond,zhou2023beyond}, we use $AP_c$, $AP_n$, and $AP_{cn}$ to represent the precision metrics for predicting the object's category, the object's nature, and their combination, respectively.
Additionally, an extra prefix 'm' is added to represent mean AP, i.e., mAP.

\begin{table}[t]
\centering
\scriptsize
\setlength\tabcolsep{2.0pt} 
\caption{The evaluation results of the five proposed baseline detectors, i.e., RO-YOLOv8, RO-YOLOv10, RO-RTMDet, RO-YOLOX, and RO-PPYOLOE, on the RODD dataset. It is important to note that $AP_c$, $AP_n$, and $AP_{cn}$ represent the precision metrics for predicting the object's category, the object's nature, and the combination of both.}
\label{table_overall_performance}
\begin{tabular}{cccc}
\toprule[1pt] 
           & \{$AP_c$, $AP_n$, $AP_{cn}$\}@0.5 & \{$AP_c$, $AP_n$, $AP_{cn}$\}@0.75 & \{$mAP_c$, $mAP_n$, $mAP_{cn}$\} \\ \hline \hline
RO-YOLOv8  & (0.693, 0.736, 0.583)  & (0.663, 0.700, 0.561)   & (0.638, 0.671, 0.540) \\
RO-YOLOv10 & (\textBF{0.766}, \textBF{0.789}, \textBF{0.585})  & (\textBF{0.726}, \textBF{0.751}, \textBF{0.562})   & (\textBF{0.697}, \textBF{0.723}, \textBF{0.542}) \\
RO-RTMDet  & (0.650, 0.706, 0.526)  & (0.620, 0.681, 0.506)   & (0.591, 0.654, 0.485) \\
RO-YOLOX   & (0.668, 0.700, 0.497)  & (0.594, 0.633, 0.447)   & (0.527, 0.558, 0.387) \\
RO-PPYOLOE & (0.579, 0.654, 0.532)  & (0.550, 0.621, 0.512)   & (0.526, 0.590, 0.486) \\ \hline 

\toprule[1pt] 
\end{tabular}
\end{table}

\subsection{Evaluation Results}
\textBF{Overall performance.}
We extensively evaluate RO-YOLOv8, RO-YOLOv10, RO-RTMDet, RO-YOLOX, and RO-PPYOLOE, the five baseline detectors proposed in this paper, on the RODD dataset.
Table \ref{table_overall_performance} presents the evaluation results using the three precision metrics defined in Section \ref{section_metrics}, i.e., $AP_c$, $AP_n$, and $AP_{cn}$.
As can be seen, RO-YOLOv10 is the best detector, consistently achieving highest Average Precisions (APs) compared to other detectors.
Besides, the evaluation results of object categories in the five baseline detectors show lower Average Precision (AP) at fixed IoUs, specifically at 0.5 and 0.75, as well as lower mean AP compared to the evaluation results of the object's nature. For all these detectors, the differences between the APs for categories and the APs for nature exceed 2\%. Notably, the largest gap is up to 7.5\%, observed in the RO-PPYOLOE detector. This indicates that identifying and localizing the object itself is more challenging than recognizing the object's nature in the images within RODD.
This discrepancy may be explained by the fact that the number of natures (i.e., 2) is significantly smaller than the number of categories (i.e., 10) in RODD. The smaller number of natures means that there is more sufficient training data for each nature class, leading to a more effective training process. In contrast, the larger number of categories results in a more dispersed dataset for each object type, making it harder for the model to learn and generalize effectively. This imbalance in the training data distribution can lead to more robust performance in recognizing the object's nature compared to identifying and localizing the object itself. This suggests that as the number of categories increases, it becomes necessary to implement effective methods that take this factor into account. 
It is noteworthy that when conventional object detection and the prediction of objects' nature are integrated into a composite task, namely detecting reflected objects, the detectors exhibit lower performance compared to handling each task independently. This observation suggests that the task of detecting reflected objects we propose in this work is more challenging than conventional object detection tasks.
This lower performance highlights the importance of developing specialized algorithms and training strategies that can better manage the intricacies of this composite task. 

\begin{figure*}[t]
	\centering
\includegraphics[width=1.0\textwidth]{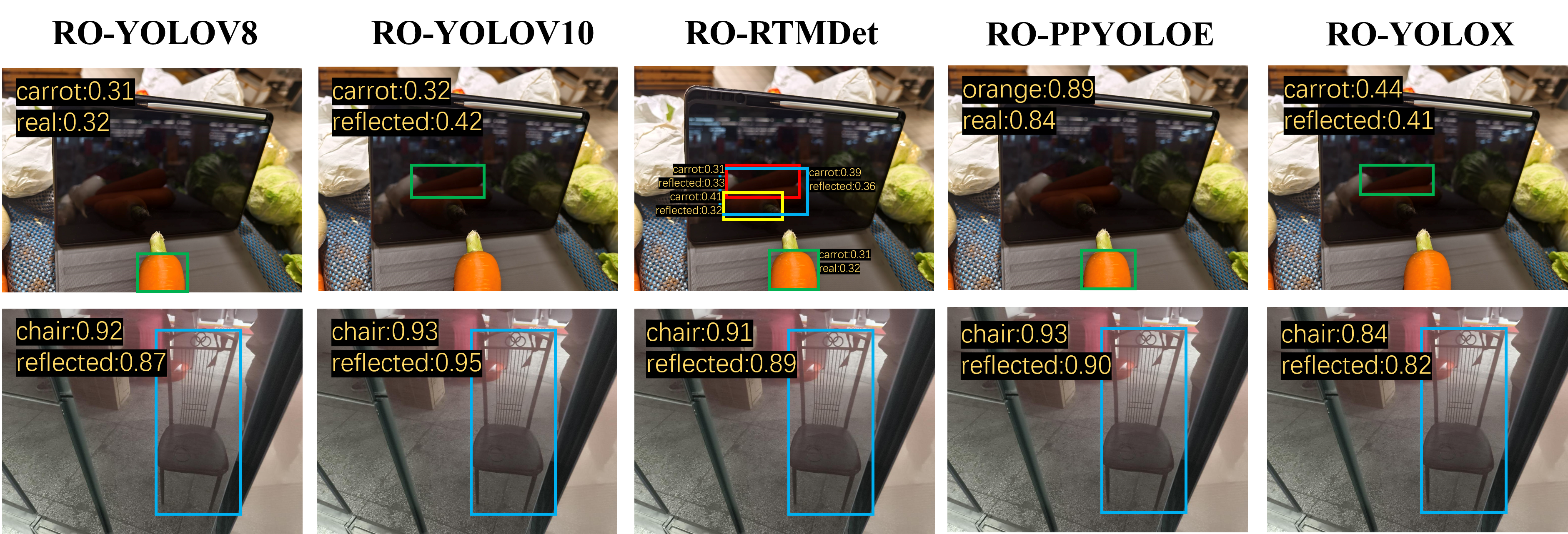}
\caption{A qualitative comparison of the five detectors on 2 samples from the carrot and the chair category, respectively. Note that all these detectors successfully detect the chair but fail to detect the carrots correctly. The predicted bounding box, object category, object nature, and the corresponding scores have been marked in the images. }
 \label{category_based_figs}
\end{figure*}
\begin{table}[h]
\centering
\caption{Comparison of the $mAP_{cn}$ of the five baseline detectors on the RODD dataset. It is important to note that $mAP_{cn}$ is the mAP for prediction of the composite of the object's category and its nature.}
\label{table_class_performance}
\scriptsize
\setlength\tabcolsep{0.5pt} 
\begin{tabular}{@{}ccccccccccc@{}}
\toprule[1pt] 
                  & bowl  & apple & mouse & keyboard & banana & carrot & cup   & orange & chair  & book  \\ \hline \hline
 $mAP_{cn}$(RO-YOLOv8)  & 0.691 & 0.561 & 0.744 & 0.780     & \textBF{0.683}  & \textBF{0.484}  & 0.787 & 0.522  & 0.837 & 0.623 \\
 $mAP_{cn}$(RO-YOLOv10) & \textBF{0.718} & \textBF{0.592} & 0.761 & 0.775    & 0.679  & 0.483  & \textBF{0.847} & \textBF{0.650}   & \textBF{0.841} & \textBF{0.629} \\
 $mAP_{cn}$(RO-RTMDet)  & 0.673 & 0.478 & \textBF{0.781} & \textBF{0.788}    & 0.645  & 0.260   & 0.736 & 0.387  & 0.829 & 0.338 \\
 $mAP_{cn}$(RO-YOLOX)   & 0.629 & 0.526 & 0.641 & 0.612    & 0.492  & 0.259  & 0.660  & \textBF{0.650}   & 0.632 & 0.554 \\
 $mAP_{cn}$(RO-PPYOLOE) & 0.690  & 0.472 & 0.763 & 0.787    & 0.622  & 0.254  & 0.752 & 0.387  & 0.825 & 0.349 \\ \hline 
\toprule[1pt] 
\end{tabular}
\end{table}

\textBF{Performance on per Category.}
To get a deeper analysis and understanding of the performance in detecting the nature of objects using our proposed baseline detectors, we further conduct performance evaluations on each category.
Table \ref{table_class_performance} presents the $mAP_{cn}$ of the five detectors evaluated on RODD.
As observed, the five detectors demonstrate their best performance on the chair category and their worst on the carrot category. The $mAP_{cn}$ values for the chair category are all above 80\%, except for the RO-YOLOX detector. In contrast, for the carrot category, the $mAP_{cn}$ values are all below 50\%, with RO-RTMDet, RO-YOLOX, and RO-PPYOLOE even falling below 30\%. This disparity could be explained by the fact that images typically contain a single chair target, often presented at standard sizes. Conversely, images frequently contain numerous carrot targets, leading to clusters and occlusions. See Fig. \ref{category_based_figs} for an intuitive comparison of examples of these two categories, the first row is the qualitative results of the carrot obtained by the five detectors.  
The challenges in detecting these carrots on the screen are compounded by their size and the properties of the reflective surfaces themselves.  The screen has a lower reflectivity coefficient compared to mirrors, which obscures the appearance of the carrots when reflected. This reduced reflectivity makes it difficult for detectors such as RO-YOLOV8 and RO-PPYOLOE to accurately identify the characteristics of carrots in their reflected versions. Moreover, carrots, being similar in color and appearance to oranges, can further confuse detectors like RO-PPYOLOE, leading to misjudgments in object categorization. In contrast, these detectors succeed in detecting the chair due to the high reflectivity of the mirror and the absence of cluttered backgrounds.
In addition, the experimental results in Table \ref{table_class_performance} also demonstrate that the same detectors will achieve varying performance across different categories. This variation may be attributed to inherent differences in object characteristics, such as size, shape, texture, and context within the images, as well as the unbalanced distribution of categories. These results underscore the importance of considering object-specific challenges in detecting reflected objects.


\begin{figure*}[t]
	\centering
\includegraphics[width=1.0\textwidth]{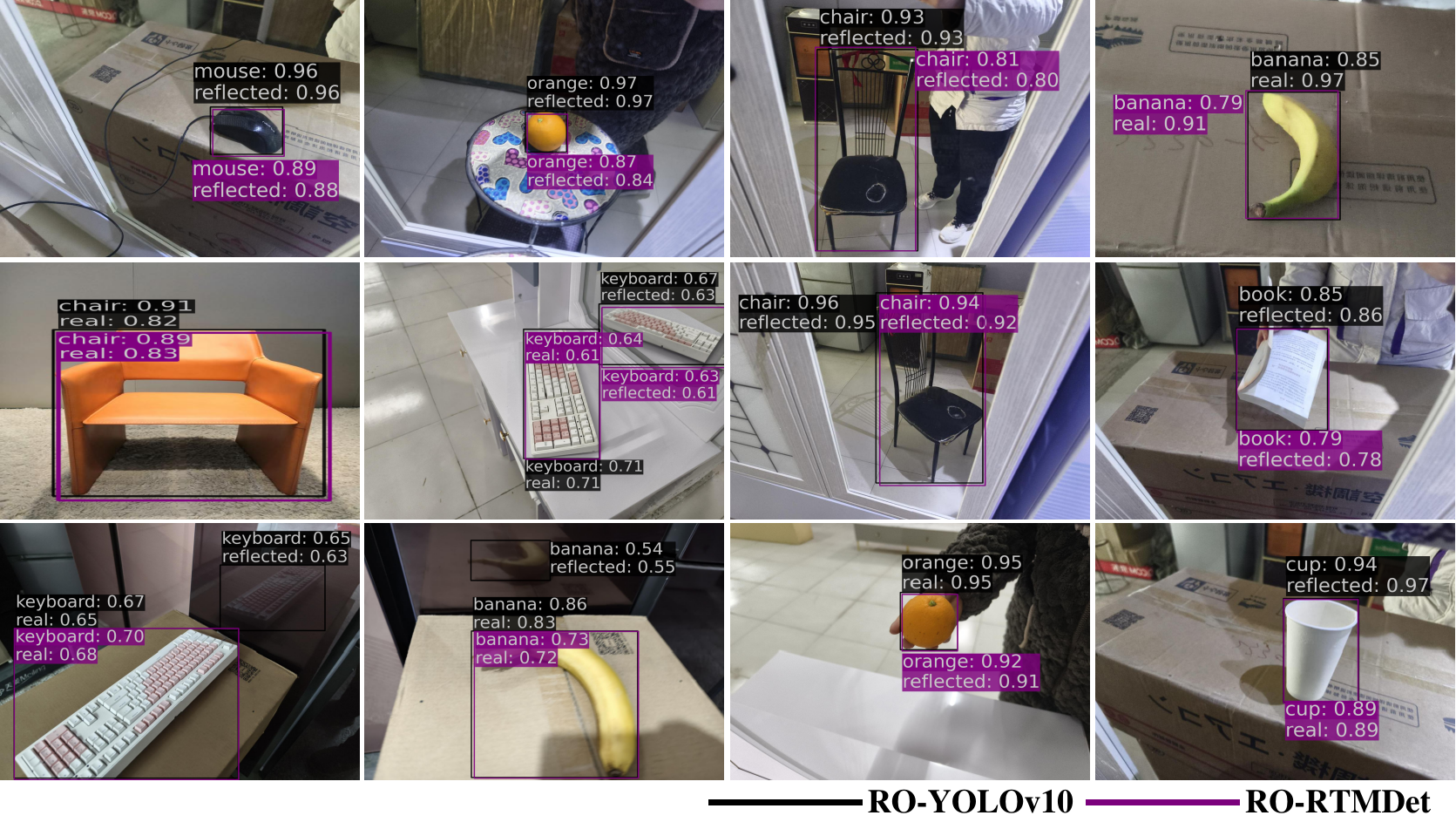}
\caption{A qualitative evaluation was conducted on 12 samples from RODD. The first two rows display examples accurately predicting the nature of objects using RO-YOLOv10 and RO-RTMDet detectors, while last row shows error detection results generated by these two detectors. Note that the predicted bounding box, object category, object nature, and the corresponding scores have been marked in the images. }
 \label{qat_figs}
\end{figure*}

\textBF{Qualitative Evaluation.}
Given the potential for overwhelming viewers with too many methods in a single image, Fig. \ref{qat_figs} presents qualitative detection results from just the RO-YOLOv10 and RO-RTMDet detectors. The first two rows display eight correctly predicted samples, while the third row shows examples where the detectors inaccurately predicted the object's nature. In these cases, reflected objects might blend into low-light backgrounds or lack distinct texture features (i.e., the first and second samples), or their mirrored background may resemble the real background (i.e., the third and fourth samples), leading to missed or inaccurate detections. This evaluation highlights that in complex scenes, the detectors are prone to struggle with accurately identifying the nature of objects.

\begin{table}[htbp!]
\centering
\scriptsize
\caption{The ablation study of the RO-YOLOv10 model is conducted on RODD using various weighting coefficients.}
	\label{tabble_loss_weighting}
 \setlength\tabcolsep{2.5pt} 
\begin{tabular}{cccc}
\toprule[1pt] 
$\lambda$ & \{$AP_c$, $AP_n$, $AP_{cn}$\}@0.5 & \{$AP_c$, $AP_n$, $AP_{cn}$\}@0.75 & \{$mAP_c$, $mAP_n$, $mAP_{cn}$\} \\ \hline \hline
0.2    & (0.744, 0.765, 0.552)  & (0.706, 0.731,  0.526)   & (0.682, 0.706, 0.514) \\
0.4    & (0.745, 0.779, 0.568)  & (0.710, 0.742, 0.549)   & (0.682, 0.716, 0.530) \\
0.6    & (0.742, 0.776, 0.553)  & (0.708, 0.738, 0.533)   & (0.679, 0.710, 0.515) \\
0.8    & (0.751, 0.776, 0.567)  & (0.716, 0.739, 0.544)   & (0.684, 0.708, 0.526) \\
1.0      & (\textBF{0.766}, 0.789, \textBF{0.585})  & (\textBF{0.726}, 0.751, \textBF{0.562})   & (\textBF{0.697}, 0.719, \textBF{0.542}) \\
1.2    & (0.749, 0.772, 0.560)  & (0.708, 0.729, 0.532)   & (0.682, 0.705, 0.518) \\
1.4    & (0.743, 0.771, 0.563)  & (0.705, 0.738, 0.540)   & (0.672, 0.705, 0.519) \\
1.6    & (0.758, \textBF{0.790}, 0.576)  & (0.721, \textBF{0.754}, 0.557)   & (0.688, \textBF{0.723}, 0.534) \\
1.8    & (0.747, 0.775, 0.524)  & (0.712, 0.739, 0.545)   & (0.680, 0.707, 0.535) \\
2.0      & (0.743, 0.784, 0.574)  & (0.707, 0.746, 0.549)   & (0.674, 0.710, 0.525) \\
\toprule[1pt] 
\end{tabular}
\end{table}

\subsection{Ablation Study}

We train the RO-YOLOv10 model on RODD using different weighting coefficients, i.e., $\lambda$ in Eq. (\ref{yolov10_loss}), which varies from 0.2 to 2.0 in steps of 0.2, in order to study the impact of the coefficient for weighting the loss of predicting the nature of objects.
This experiment aims to determine how different weightings influence the model's ability to balance the two tasks: detecting the objects and predicting their nature. By adjusting $\lambda$, we can observe how the model prioritizes the nature prediction task relative to the conventional object detection task.
Table \ref{tabble_loss_weighting} presents the experimental results for the mAP and the AP at fixed IoUs (0.5 and 0.75).
As can be seen, RO-YOLOv10 can obtain part of the best APs when $\lambda$ is set to 1.0 or 1.6.
An obvious partial difference between $AP_c$ and $AP_n$ is evident.
In general, as $\lambda$ ranges from 0.2 to 1.6, $AP_c$ initially increases and then decreases, while $AP_n$ consistently rises, reaching their optimal values at 1.0 and 1.6, respectively.
A compromise is achieved when $\lambda$ is set to 1.0, serving as the default setting, where $AP_{cn}$ reaches its highest value.
Specifically, the highest $mAP_c$, which equals 0.697, occurs at $\lambda$ = 1.0; the highest $mAP_n$, which equals 0.723, occurs at $\lambda$ = 1.6; and the highest $mAP_{cn}$, which equals 0.542, occurs at $\lambda$ = 1.0.
The results suggest that there may be a counteracting impact between object localization and prediction of objects' nature when these two tasks are done concurrently as a composite task.
More effective methods for mitigating this counteracting effect are needed.

\section{Conclusions}

In this paper, we investigated the underexplored challenge of reflective object detection and introduced the Reflective Object Detection Dataset (RODD), an extensive benchmark specifically designed for this task. RODD includes 10 categories and 21,059 images of real or reflected objects in various backgrounds, accompanied by standard annotations of bounding boxes and the nature of the objects (real or reflected), distinguishing it from traditional object detection benchmarks. In addition to introducing RODD, we adapted five state-of-the-art object detection models to this challenging task and presented baseline results. The experimental findings reveal the limitations of current methods when applied to reflected object detection, underscoring the necessity for specialized approaches. By releasing RODD, we aim to foster and advance future research in detecting reflected objects. This dataset provides a valuable resource for developing and evaluating new methods, ultimately contributing to improved performance in applications such as surveillance, autonomous driving, service robots, smart homes, and medical imaging. 


%
%


\end{document}